\documentclass[runningheads]{llncs}
\usepackage{graphicx}
\usepackage{hyperref}
\usepackage{booktabs}
\usepackage{array}
\usepackage{multirow}
\usepackage{siunitx}
\usepackage{enumitem}
\begin{document}
\title{Prediction of Clinical Complication Onset using Neural Point Processes}
%
%


\author{Sachini Weerasekara \and
Sagar Kamarthi \and
Jacqueline Isaacs}
%
%
\institute{Northeastern University, 360 Huntington Ave, Boston, MA 02115, USA}
%
\maketitle              
\begin{abstract}

Predicting medical events in advance within critical care settings is paramount for patient outcomes and resource management. Utilizing predictive models, healthcare providers can anticipate issues such as cardiac arrest, sepsis, or respiratory failure before they manifest. Recently, there has been a surge in research focusing on forecasting adverse medical event onsets prior to clinical manifestation using machine learning. However, while these models provide temporal prognostic predictions for the occurrence of a specific adverse event of interest within defined time intervals, their interpretability often remains a challenge. In this work, we explore the applicability of neural temporal point processes in the context of adverse event onset prediction, with the aim of explaining clinical pathways and providing interpretable insights. Our experiments span six state-of-the-art neural point processes and six critical care datasets, each focusing on the onset of distinct adverse events. This work represents a novel application class of neural temporal point processes in event prediction. 


\end{abstract}
\section{Introduction}

Timely interventions in critical care significantly impact patient outcomes, particularly in responding to adverse clinical events like cardiac arrest, pneumonia, and sepsis. However, predicting the onset of these events before they clinically manifest remains challenging due to the intricate clinical presentation of critically ill patients. Machine learning models offer predictive capabilities by capturing complex correlations within the dynamic patient conditions over time, assisting medical professionals in effectively anticipating the onset of such critical events \cite{bakar2023review,chou2023predicting,moor2021early,natu2022review,song2021comparison}. 

In particular, these models employed various neural architectures, including recurrent architectures such as long-short-term memory (LSTM) \cite{hochreiter1997long}, gated recurrent units (GRU) \cite{cho2014learning}, convolutional networks \cite{Fukushima1983}, and temporal convolution networks \cite{Lea2017,Oord2016}. Moreover, several studies implemented boosted tree-based models, including XGBoost \cite{Chen2016} and random forest \cite{Kam1995}. These models commonly utilize feature spaces, including a variety of patient biomarkers, encompassing vital signals, lab measurements, and demographic data (Moor et al., 2021). 

While the predictive prowess of these models continues to advance, many state-of-the-art (SOTA) models offer probabilities for disease onset within a given timeframe but often fall short in terms of interpretability regarding these probabilities. Research on interpretable predictions has predominantly employed attention mechanisms \cite{attentionisallyouneed} to discern feature importance \cite{Alsaleh2023}. Nonetheless, addressing interpretability still remains challenging, given the inherently complex nature of neural architectures compounded by the high dimensional nature of patient biomarkers. 

By shifting the problem's perspective to predicting a series of continuous-time events leading up to the occurrence of the adverse event of interest, there's potential for a substantial enhancement in the interpretability of predictions regarding adverse event occurrence. This involves employing temporal point process (TPP) modeling to predict a sequence of continuous-time events.

Temporal point processes serve as a powerful framework for understanding a series of events unfolding temporally. They allow capturing the timing, frequency, and dependencies of events in continuous time. More specifically, TPPs describe event timing, probability of occurrence, and temporal dependencies for a series of forthcoming events. Classical TPPs, such as Poisson processes \cite{classicalpoisontpp} and Hawkes processes \cite{classicalhawkestpp}, have long served as foundational mathematical models and found extensive application on event series predictions in diverse domains such as traffic modeling \cite{cramertraffic}, finance \cite{Hasbrouckfinance}, and seismology \cite{Ogataseismology}. Nevertheless, the strong parametric assumptions inherent in these models tend to limit their ability to capture the intricate complexities of real-world phenomena. 

\begin{figure}[tbh]
\centering
  \includegraphics[width=100mm]{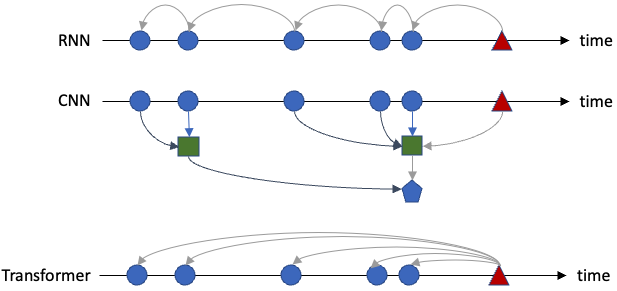} 
  \caption{Illustration of dependency calculation in RNN, CNN, and transformer-based models with the current event (red triangle) and historical events (blue circles).}
  \label{Fig 4}
\end{figure} 

Researchers have recently explored neural TPPs to address these limitations, which leverage neural networks' expressiveness to learn high-order dependencies \cite{Weerasekara2021Thesis,Weerasekara2022Trends,Weerasekara2024RL,Weerasekara2025Improvements}. Because of the sequential nature of event streams, the majority of the neural TPPs employ recurrent neural networks (RNNs), as RNNs have representation power to model the dynamics of sequence data. Some studies also explore convolutional neural networks \cite{convtppoord,convtppgehring,convtppyin} that are tailored to analyze sequential data. More recently, neural TPPs have deployed transformer architectures based on self-attention modules and have been proven superior to RNN or CNN-based models \cite{ZuoTHP}. Modules in transformer architectures model dependencies among events that are at any temporal distance from each other by assigning attention scores. This has enabled transformer-based models to capture both short-term and long-term dependencies. Figure 1 illustrates dependency computations of RNNs, CNNs, and transformer-based models. 

In this study, we employ six state-of-the-art (SOTA) neural TPPs to investigate their potential for interpretable adverse event prediction in critical care settings. To the best of our knowledge, this study represents the first attempt to apply neural point processes for predicting the onset of adverse clinical events. Our primary contributions include:

\begin{itemize}[label=\textbullet]
    \item We frame the adverse event onset prediction problem within the framework of neural temporal point processes.
    \item We employ six SOTA neural TPP models to predict six adverse events-pneumonia, sepsis, cardiac arrest, acute renal failure, respiratory failure, and cardiogenic shock.
    \item Analyzing six critical care event onset prediction scenarios, we demonstrate the extent of effectiveness of current SOTA methodologies under a defined standard and discuss avenues for further improvement. 
    
\end{itemize}

\section{Method}

\subsection{Continuous-time Event Prediction}

Consider a fixed time duration $[0,T]$ during which we observe an event sequence. Let $N$ denote the total number of events occurring within this interval, happening at times $0<t_1<....<t_N$. This event sequence can be represented as $(t_1,e_1),....(t_n,e_N)$, where $e_k\in{1,....K}$ represents $K$ distinct event types. If we denote the probability of an event of type $k$ occurring within a time interval $[t, t+dt)$ as $p_k(t|x_{[0,t)})$, then the probability of no event occurring during this interval would be $1-\sum_{k=1}^K p_k(t|x_{[0,t)})$. The distribution of a TPP is defined by the intensity function $\lambda_k(t|x_{[0,t)})\geq0$ for each event type $k$ at each time $t>0$ such that $p_k(t|x_{[0,t)}) = \lambda_k(t|x_{[0,t)})dt$. Hence, the next event in the sequence is determined by the maximum event intensity $argmax(\lambda_k(t|x_{[0,t)})dt)$.

\begin{figure}[tbh]
\centering
  \includegraphics[width=70mm]{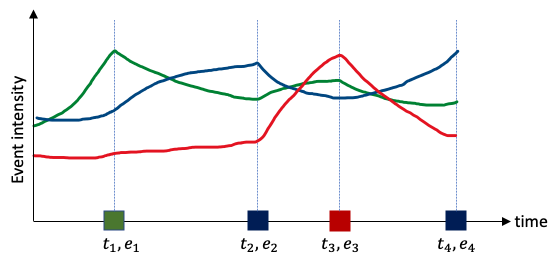} 
  \caption{Illustration of determining an event stream from a neural TPP. Each event intensity function is a continuous parametric curve determined by the observed history of event occurrences. Event intensity curves kee updating as events occur.}
  \label{Fig 4}
\end{figure} 

Neural TPPs autoregressively generate event sequences determined by neural networks. For the $i-th$ event, it computes the embedding of the event $e_i \in \in R^D$ via an embedding layer, and the hidden state $h_i$ gets updated conditioned on $e_i$ and the previous state $h_{i-1}$. Then the prediction for the next event conditioned on the hidden state $h_i$ is drawn as below where $f$ denotes a recurrent encoder, which is either RNN-based or attention-based.

\begin{equation}
t_{i+1},e_{i+1} \sim P_\theta(t_{i+1},e_{i+1}|h_i), h_i = f(h_{i-1},e_i)
\end{equation}

 Both classical TPPs and neural TPPs use negative log-likelihood (NLL) as the training loss function. The NLL of a TPP given the event sequence $e_{[0,T]}$ is given in Equation (2), and the derivation can be found elsewhere \cite{ZuoTHP}.

\begin{equation}
\sum_{i=1}^Nlog\lambda_{k_i}(t_i|e[0,t_i)) - \int_{t=0}^T\sum_{k=1}^K\lambda_k(t|e_{[0,t)})dt \hspace{5mm}
\end{equation}

\subsection{Adverse Event Onset with TPP}

Next, we formulate the adverse event prediction problem in the context of TPP. Suppose we are given an event sequence $S=\{(t_i,e_i)\}_{i=1}^I$ of I events where each event $e_i \in \{adverse,amber-flag\}$ and $e_i \in \{1,...,K\}$ where K is the number of different event types from an event pool with one adverse event and several amber flag events. An amber flag event describes an abnormal reading of patient biomarkers: vital signals or lab measurements. For instance, abnormal body temperature is translated as an amber flag event named "thermoregulation dysfunction". Figure 3 illustrates an amber flag event sequence that occurred within 24 hours of a cardiac arrest onset.

\begin{figure}[tbh]
\centering
  \includegraphics[width=100mm]{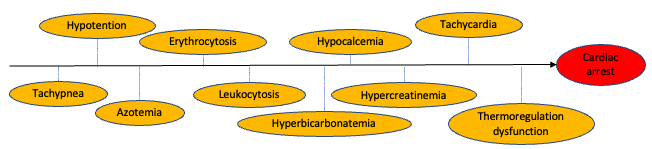} 
  \caption{A sample series of amber flag events (orange) preceding the onset of the adverse event (red) within 24 hours of the adverse event onset.}
  \label{Fig 4}
\end{figure} 

\textbf{Data preprocessing.} Following the standard practices, as the first step, we split the data into train, test, and validation sets. Then, to feed the model with sequences of varying lengths, we pad the rear end of the event sequences to the same length and use a masking tensor to identify padding event tokens. Whenever attention-based TPPs were implemented, an attention-mask was employed to avoid looking at events in the future. 

\textbf{Training} We implemented six SOTA models using Pytorch, as discussed in section 3. We optimize the model parameters by locally maximizing log-likelihood in Equation (2) using stochastic gradient descent. Since the integral in the second term of the loss function makes it challenging, we approximate the integral by Monte-Carlo estimation to compute the overall log-likelihood. 

\textbf{Inference.} With the learned parameters, we apply an MLP layer at the end of each network to predict the event type and time.

\section{Experiments}

In this section, we discuss the experiment setup: datasets, model implementation and assessment, and result analysis. 

\subsection{Setup}

\textbf{Model implementation.} We conduct comprehensive evaluations of the following six widely cited SOTA neural point processes that facilitate multi-type event predictions regarding their effectiveness in addressing the adverse event onset problem. Of these, three utilize Recurrent Neural Network (RNN) architecture, while the remaining three leverage attention-based mechanisms.

\begin{itemize}[label=\textbullet]
    \item Three RNN-based models: Neural Hawkes Process (NHP) \cite{MeiRNN}, Recurrent Marked Temporal Point Process (RMTPP) \cite{DuRNN}, Intensity-free \cite{Intensityfree}
    \item Three attention-based models: Transformer Hawkes Process (THP) \cite{ZuoTHP}, Self-attentive Hawkes Process (SAHP) \cite{Zhangsahp}, Attentive Neural Hawkes Process (A-NHP) \cite{YangAttNHP}
\end{itemize}

To ensure consistency in implementation standards, all models were developed in Pytorch, and we employed base models provided by EasyTPP \cite{EasyTPP} with the default hyperparameters for our experiments. However, instead of relying on the thinning algorithm utilized in EasyTPP, we opted to integrate a Multilayer Perceptron (MLP) layer for event type and time prediction. Our code implementation can be accessed at \href{https://github.com/sachininw/NeurlTPP-for-Adverse-Event-Onset-Prediction}{NTPP-for-clinical-events}. 

\textbf{Training objective.} For the training objective, which is log-likelihood, we used the implementation given in EasyTPP. The integral in the second term is computed using the Monte Carlo approximation given by Mei \& Eisner \cite{MeiRNN}.

\textbf{Next event type and time prediction.} For each event in the validation dataset, we try to predict the immediate next event $e_{i+1}$ and the time at which the event is predicted to happen $t_{i+1}$. For all the intensity-based models, we use an MLP layer with a softmax function at the end of the network to compute the next event time $\int_{t_i}^\infty t.p(t|h_i)$ and event intensities $\lambda_e((t_j+1)|h_i)$. For the IF model, we use the conditional distribution of inter-event times to determine the next event time and type as implemented in EasyTPP. 

\textbf{Datasets.} We conducted experiments using five adverse event datasets obtained from the eICU Collaborative Research Database \cite{eICU}. This database is a comprehensive collection of anonymized patient data from over 200,000 ICU admissions across various medical centers in the United States between 2014 and 2015. It encompasses a wide range of information, including patient demographics, vital signs, laboratory measurements, and diagnostic records.

Event chains were constructed using vital signs, laboratory measurements, and diagnosis records. Abnormal readings of patient biomarkers served as indicators for the onset of amber flag events, as elaborated in Section 2. The onset of adverse events was determined from diagnostic details.

The five datasets comprise time-stamped positive samples, representing event sequences of patients who experienced the respective adverse event within 12 hours of onset, alongside negative samples, which correspond to event sequences of patients who did not encounter the adverse event. After pre-processing, the below positive event sequences are drawn from a cohort of 17014 pneumonia patients, 20256 sepsis patients, 4945 cardiac arrest patients, 17005 acute renal failure patients, 33054 respiratory failure patients, and 1621 cardiogenic shock patients.

Table 1 presents statics of the datasets considered in the study.

\begin{table}[htb]
\centering
\caption{Dataset statistics. Negative samples are downsampled to achieve a balanced dataset, aligning the number of negative samples with that of positive samples due to the significantly low positive-to-negative ratio. Consequently, each dataset comprises an even distribution of positive and negative samples. $K$ is the total number of selected amber flag events, which is the same across all the complications.}
\newcolumntype{C}{>{\leftalign\arraybackslash}X}
\resizebox{\textwidth}{!}
{\begin{tabular}{@{}rllllllll@{}} 
\toprule

\multirow{2}{*}{\textbf{Dataset}} & 
\multirow{2}{*}{\textbf{$K$}} &
\multicolumn{3}{c}{\textbf{\# of event sequences (positive/negative)}} & 
\multicolumn{3}{c}{\textbf{Sequence length}}\\[0.3cm] 
 & & \textbf{Train} & \textbf{Test} & \textbf{Dev} & \textbf{Min} & \textbf{Mean} & \textbf{Max}\\ [0.3cm] \midrule
Pneumonia & 34 & 
(11,910/14,000) & (2,552/3,000) & (2,552/3,000) & 2 & 14 & 28\\ [0.3cm] 
Sepsis & 34 & 
(14,179/14,000) & (3,038/3,000) & (3,039/3,000) & 2 & 12 & 29\\ [0.3cm] 
Cardiac arrest & 34 & 
(3,462/3,500) & (742/750) & (741/750) & 2 & 15 & 28\\ [0.3cm] 
Acute renal failure & 34 & 
(11,904/14,000) & (2,551/3,000) & (2,550/3,000) & 2 & 14 & 30\\ [0.3cm] 
Respiratory failure & 34 & 
(23,138/21,000) & (4,958/4,500) & (4,958/4,500) & 2 & 13 & 28\\ [0.3cm] 
Cardiogenic shock & 34 & 
(1,135/1,190) & (243/255) & (243/255) & 2 & 15 & 28\\ [0.3cm] 
\hline
\hline
\end{tabular}}
\end {table}

\textbf{Assessment Framework.} To ensure a fair comparison, we standardized the training procedure across all models: employing the Adam optimizer \cite{adamopt} with default parameters, initializing biases to zeros, omitting learning rate decay, and defining a uniform maximum number of training epochs.

The following three key questions and corresponding solution approaches guide the evaluation.

\vspace{10pt}

\noindent \textbf{Q1.} What is the Goodness-of-fit of predicted sequences? \\We train the model parameters using the training set and evaluate the log-likelihood of predicting event sequences on the validation set.

\noindent \textbf{Q2.} What is the next event prediction accuracy? \\We assess the model's ability to accurately predict the immediate next event on the event sequences in the validation set by checking if $\hat{e}_{i+1}$ and $\hat{t}_{i+1}$ are exactly similar to $e_{i+1}$ and $t_{i+1}$. Our aim is to test the performance on a standard setting, as described in section 2, rather than optimizing each model's performance. 

\vspace{5pt}

\noindent \textbf{Q3.} How far ahead can the models forecast into the future? \\ We define two event horizons—long-term, predicting event occurrences within the subsequent 12 hours (mean event count of 6), and short-term, predicting events within the ensuing 6 hours (mean event count of 3). We evaluate the models based on the similarity between event predictions in the future horizon $\hat{e}_{(t,t+T]}$ and the ground truth $e_{(t,t+T]}$ using optimal transport distance (OTD) \cite{OTD}. OTD measures the cost of aligning the predicted sequence with the ground truth sequence \cite{EasyTPP}.

\vspace{5pt}


\subsection{Results and Analysis}

\begin{figure}

\begin{tabular}{ccc}
  \includegraphics[width=40mm]{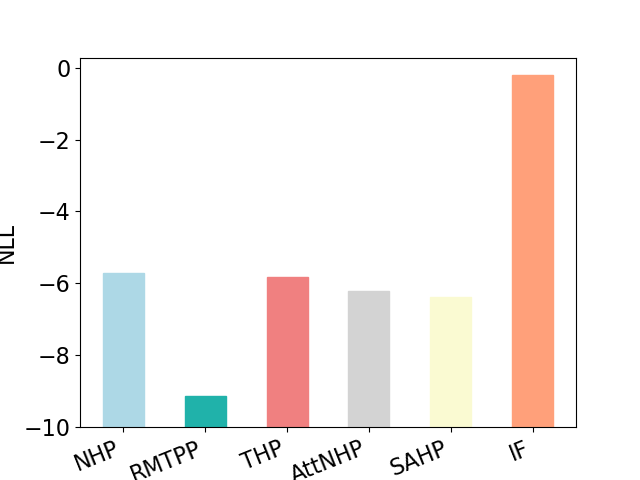} &
  \includegraphics[width=40mm]{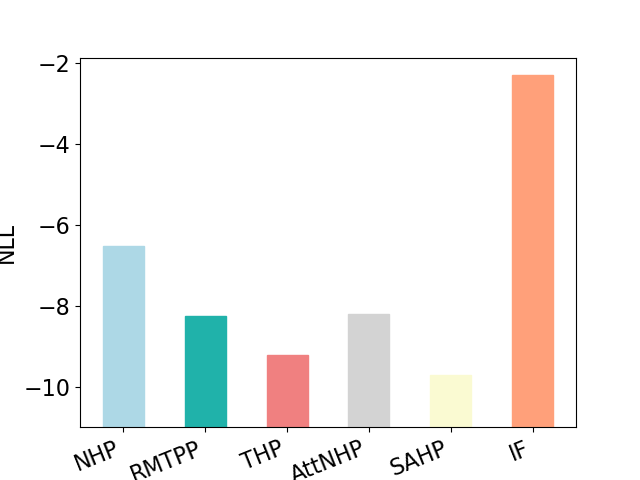} &
  \includegraphics[width=40mm]{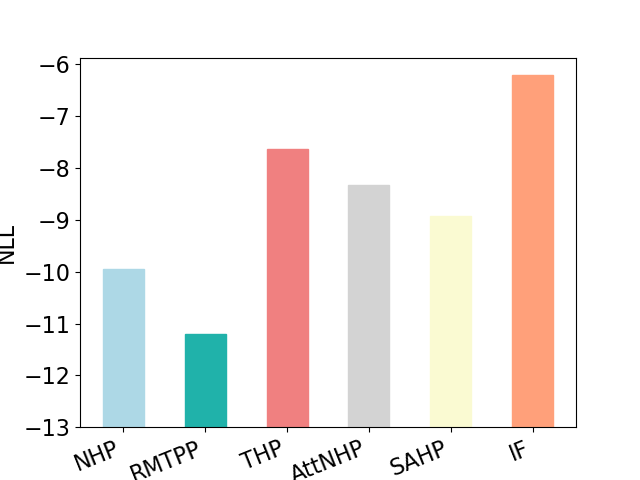} \\
  \footnotesize{Pneumonia} &
  \footnotesize{Sepsis} & 
  \footnotesize{Cardiac arrest}

\end{tabular}

\begin{tabular}{ccc}
  \includegraphics[width=40mm]{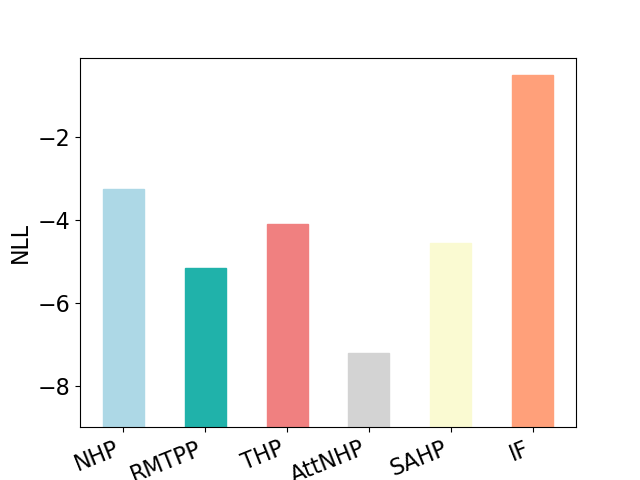} &   \includegraphics[width=40mm]{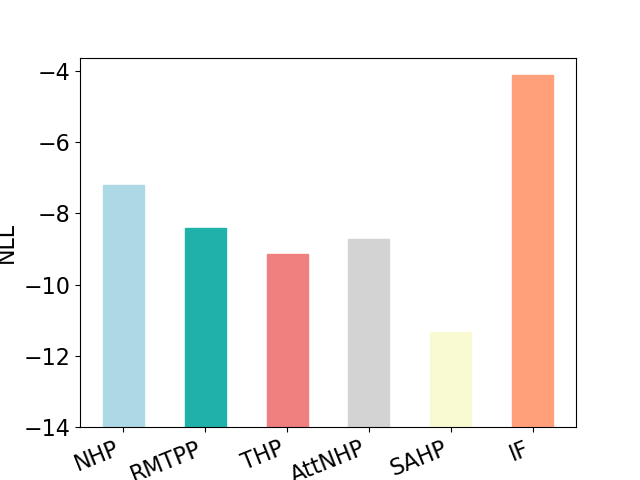}&
  \includegraphics[width=40mm]{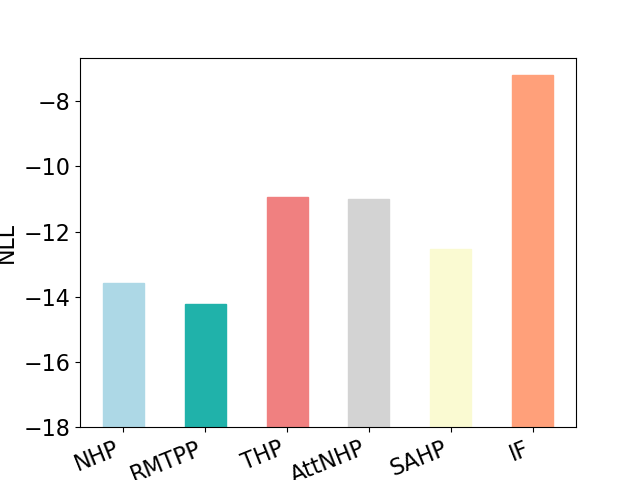}  \\
\footnotesize{Respiratory failure} & \footnotesize{Acute renal failure} & \footnotesize{Cardiogenic shock}\\

\end{tabular}
\caption{Log-likelihoods of validation dataset. Higher log-likelihoods indicate better performance.}

\end{figure}

In this section, we discuss the results of the experiments conducted with the aim of addressing the key research questions. 

Figure 4 elucidates Q1 by showcasing the negative log-likelihoods of the hold-out validation sets following a training duration of 100 epochs. Elevated values denote superior models, with IF consistently outperforming other intensity-based models across all scenarios because Monte Carlo integration in intensity-based models is prone to numerical approximation errors. 

\begin{table}[htb]
\centering
\caption{Next event type prediction performance of the six NPPs on the six datasets considered after 100 training epochs under standard assessment protocols described in section 2. We report the accuracy of the next event prediction. Higher values indicate better performance.}

\newcolumntype{C}{>{\leftalign\arraybackslash}X}
\resizebox{\textwidth}{!}
{\begin{tabular}{@{}rllllllll@{}} 
\toprule

\multirow{2}{*}{\textbf{Model}} & 
\multicolumn{6}{c}{\textbf{Next event type accuracy rate}}\\[0.3cm] 
 & Pneumonia & Sepsis & Cardiac arrest & Acute renal failure & Respiratory failure & Cardiogenic shock\\ [0.3cm] \midrule
NHP &35\%	& 34\%	&22\%	&33\%	&39\%	&26\% \\ [0.3cm] 
RMTPP &29\%	&31\%	&21\%	&31\%	&33\%	&24\% \\[0.3cm] 
THP &32\%	&32\%	&23\%	&26\%	&37\%	&25\% \\[0.3cm] 
SAHP &30\%	&31\%	&22\%	&24\%	&29\%	&21\% \\ [0.3cm] 
A-NHP &32\%	&33\%	&22\%	&20\%	&32\%	&22\% \\ [0.3cm] 
IF &33\%	&37\%	&31\%&	29\%	&41\%	&24\%\\ [0.3cm] 
\hline
\hline
\end{tabular}}
\end {table}

Addressing Q2, Table 2 presents the accuracy of predicting the next event type, from event pools of 34 events each. Our aim here is to test all the models with a standard setting, as we discussed in Section 2, rather than optimizing each model's performance. In the standard setting, no model shows significantly better performance than the others, and the performance is seen to saturate among the models. While generally attention-based models perform superior to other models, results show that RNN-based models also make strong competitors to attention-based models in this case.

Regarding Q3, Figure 5 illustrates the optimal transport distances (OTD) for short-term horizon (next 3 events) and long-term horizon (next 6 events) prediction for pneumonia, sepsis, and respiratory failure. OTDs were calculated using the function $wsasserstein\_distance$ from the $scipy$ package in Python. OTD explains the cost required to transform the predicted sequence into ground truth. Therefore, lower values imply better model performance. Short-term horizon prediction performance is observed to be higher than the long-term horizon prediction, as expected across all the combinations.  Although NHP and IF are observed to be performing better, once again, a definitive winner remains elusive among the six models.

\begin{figure}

\begin{tabular}{ccc}
  \includegraphics[width=45mm]{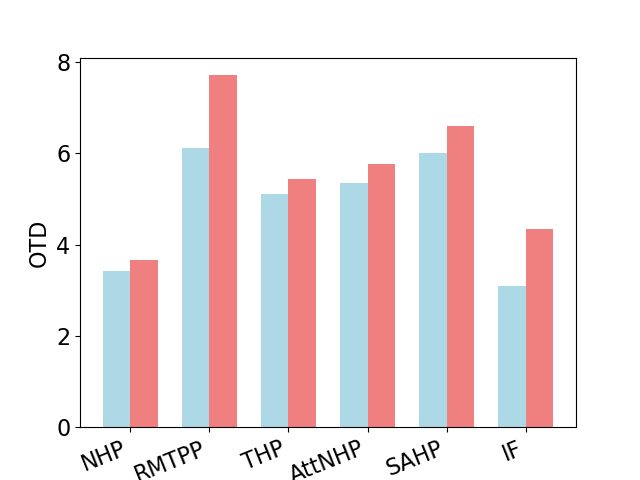} &
  \includegraphics[width=45mm]{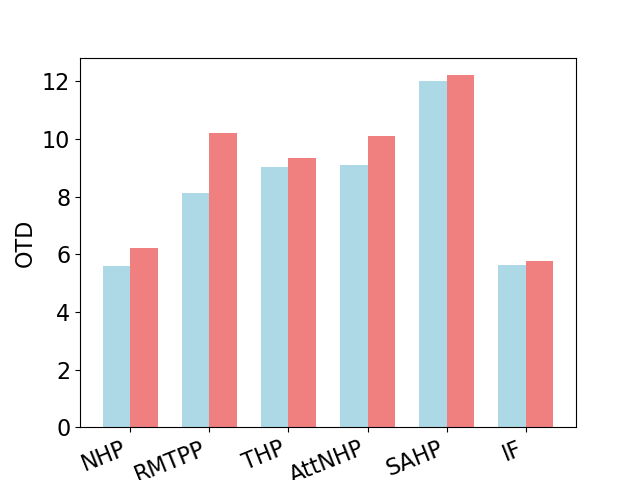} &
  \includegraphics[width=45mm]{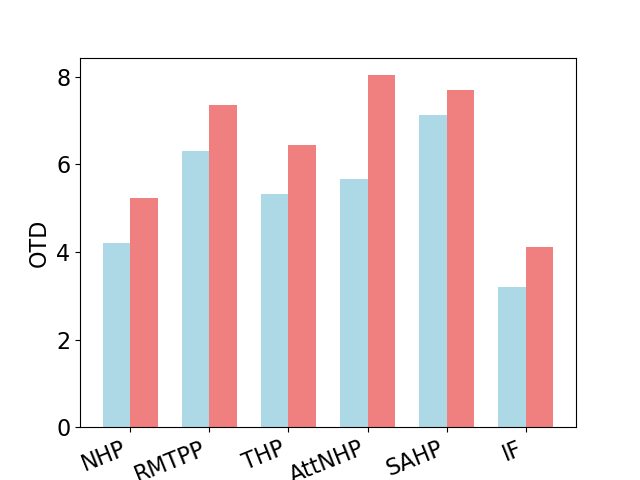} \\
  \footnotesize{Pneumonia} &
  \footnotesize{Sepsis} & 
  \footnotesize{Respiratory failure}

\end{tabular}
\caption{optimal transport distances (OTD), measuring the cost of transforming the predicted sequence to the ground truth. Low OTD values indicate better predictions.}
\end{figure} 


\section{Future Research}

In this section, we explore the prospective avenues for future research stemming from the observed results.

This study serves as an inaugural experimental investigation utilizing the foundational NTPP models in adverse clinical event prediction, thereby opening up multiple new avenues for further exploration. A key observation is the convergence of performance across diverse and sophisticated model architectures, suggesting a saturation at comparable levels. This suggests a potential enhancement through richer input feature spaces, particularly given the coarse-grained, one-hot encoded nature of input sequences for NTPP models. Hence, an avenue of future research could be the development of models capable of incorporating patient-specific details, such as demographics and biomarker values, alongside event sequences. Furthermore, since certain event sequences may be infrequent because of the complicated clinical picture of critical care patients, exploring few-shot learning strategies within this context could prove advantageous.

Generating more comprehensive and informed amber flag event sequences tailored to the specific adverse event under consideration with clinician inputs would immensely benefit event sequence predictions preceding the adverse event, thereby presenting yet another promising avenue for future research.

\section{Related work}

\textbf{Neural Temporal Point Processes (NTPP).} With the ability of deep learning models to handle sequential data, various architectures of neural point processes have emerged in recent years. Most of these models implement RNNs \cite{MeiRNN,FullyNN,Intensityfree,DuRNN} or attention mechanisms \cite{ZuoTHP,Zhangsahp,YangAttNHP} as their base architecture. Compared to RNN-based models, attention-based models have shown better performance in applications that benefit from long-term dependency computation. Compared to classical temporal point process models such as the Hawkes process \cite{classicalhawkestpp} or Poisson processes \cite{classicalpoisontpp}, these neural models show superior performance because of their ability to handle continuous state spaces and by offering greater flexibility in modeling state transition functions. With recent rapid advancements in NTPP model architectures, their applications are beginning to gain traction in various domains.

\vspace{10pt}

\noindent\textbf{Interpretable clinical event onset prediction.} Attention mechanisms \cite{attentionisallyouneed} shine in their ability to elucidate a model's decisions by emphasizing the features to which the model allocates the most attention. They have also proven invaluable in the realm of clinical event onset prediction. Consequently, a majority of interpretable clinical event onset prediction models leverage attention to determine feature importance \cite{int1,int2,int3,int4,int5}. While NTPPs can be a powerful tool in addressing interpretability by predicting events preceding the event of interest, their effectiveness within clinical contexts remains largely unexplored.

\section{Conclusion}

In this study, we present a novel perspective on predicting adverse clinical event onset. We leverage neural point processes to enhance the explainability of predictions regarding adverse event occurrences. Through experimentation with six state-of-the-art neural temporal point process models and six selected adverse events, our findings illuminate numerous promising avenues for future research. Moreover, our results underscore the potential benefits of enriching event sequence predictions by integrating diverse data sources and types to create richer feature spaces alongside historical event sequences.

%
%
%
%

\end{document}